\documentclass{article}

\usepackage{arxiv}

\usepackage[utf8]{inputenc} 
\usepackage[T1]{fontenc}    
\usepackage{hyperref}       
\usepackage{url}            
\usepackage{booktabs}       
\usepackage{amsfonts}       
\usepackage{nicefrac}       
\usepackage{microtype}      
\usepackage{graphicx}

\usepackage{float}

\usepackage{amsmath}
\usepackage{amssymb}
\newcommand{\vect}[1]{\boldsymbol{#1}} 
\everymath{\displaystyle}              

\title{Multi-task Learning for Sparse Traffic Forecasting}

\author{
 Jiezhang Li, Junjun Li, Yue-Jiao Gong\textsuperscript{*} \\
  School of Coumpute Science and Engineering, South China University of Technology \\
  Guangzhou, China \\
  \texttt{gongyuejiao@gmail.com} \\
}

\begin{document}
\maketitle
\begin{abstract}
Accurate traffic prediction is crucial to improve the performance of intelligent transportation systems. Previous traffic prediction tasks mainly focus on small and non-isolated traffic subsystems, while the Traffic4cast 2022 competition is dedicated to exploring the traffic state dynamics of entire cities. Given one hour of sparse loop count data only, the task is to predict the congestion classes for all road segments and the expected times of arrival along super-segments 15 minutes into the future. The sparsity of loop counter data and highly uncertain real-time traffic conditions make the competition challenging. For this reason, we propose a multi-task learning network that can simultaneously predict the congestion classes and the speed of each road segment. Specifically, we use clustering and neural network methods to learn the dynamic features of loop counter data. Then, we construct a graph with road segments as nodes and capture the spatial dependence between road segments based on a Graph Neural Network. Finally, we learn three measures, namely the congestion class, the speed value and the volume class, simultaneously through a multi-task learning module. For the extended competition, we use the predicted speeds to calculate the expected times of arrival along super-segments. Our method achieved excellent results on the dataset provided by the Traffic4cast Competition 2022, source code is available at https://github.com/OctopusLi/NeurIPS2022-traffic4cast. 
\end{abstract}


\section{Introduction}
Intelligent transportation system plays an increasingly important role in modern cities. As an important task of intelligent transportation systems, traffic prediction aims to predict future traffic states according to historical traffic data and real-time traffic conditions. 

In the past decade, deep learning methods have made breakthroughs in the field of traffic prediction. Some methods~\cite{ma2017learning,polson2017deep,zhene2018deep,zhang2017fcn,ke2017short,wei2019autoencoder} used Convolutional Neural Network and Recurrent Neural Network~\cite{mikolov2010recurrent} to capture the temporal dependence of traffic data. But these methods do not consider the topological relationship of traffic data. Some works~\cite{zheng2020gman,huang2020lsgcn,guo2021learning,chen2020multi} introduced Graph Neural Network~\cite{scarselli2008graph} to model road graphs and learn the spatial dependence between road segments.
However, most of the previous methods are built upon small subsystems, and meanwhile they require dense traffic data acquisition.  These methods are not applicable in the practical large-scale traffic systems with sparse measurement data. 

During the previous traffic4cast competition in 2019-2021~\cite{pmlr-v123-kreil20a, pmlr-v133-kopp21a,pmlr-v176-eichenberger22a}, many methods have been proposed to predict traffic conditions by using large-scale data and have contributed both methodological and practical insights to advance the application of AI in forecasting traffic and other spatial processes. The Traffic4cast 2022 challenge further aims to explore the ability to generalize loosely related temporal vertex data on just a few nodes to predict dynamic future traffic states on the edges of the entire road graph. 
Specifically, the core challenge of Traffic4cast 2022 is to predict the congestion classes of all road segments 15 minutes into the future. The extended challenge aims to predict the expected times of arrival along super-segments over the next 15 minutes. Note that the input of competition is the car count data from spatially sparse vehicle counters in three cities in 15-minute aggregated time bins for one hour before the prediction time slot.

\section{Methods}
Our model is composed of several modules, which are presented one-by-one in this section. The overview of the model is shown in Figure 1.

\begin{figure}[t]
	\centerline{\includegraphics[width=0.8\linewidth]{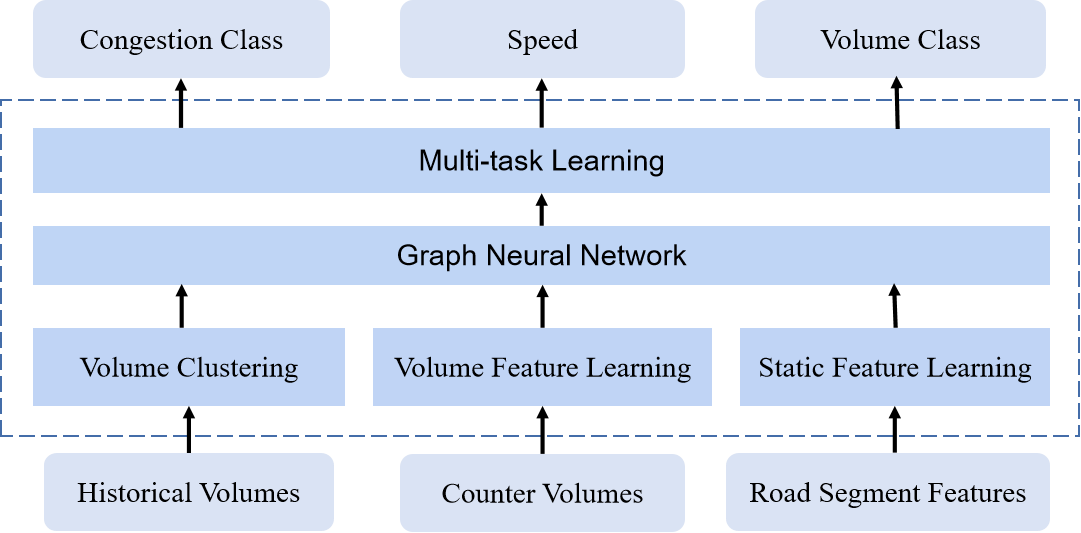}}
	\caption{ Architecture of the proposed model.}
	\label{fig:real-1}
\end{figure}

\subsection{Volume Clustering}
First, the data records are clustered into 10 groups following the simple median clustering method\footnote{https://github.com/iarai/NeurIPS2022-traffic4cast/} provided by the organizer\footnote{https://www.iarai.ac.at/traffic4cast/}. Specifically, for each data record, we add the values of all loop counter volumes to obtain a \textit{volumeSum} value.  Then, we first sort the dataset according to \textit{volumeSum} and then patition the records into $10$ equal-frequency bins. The bin number indicates the cluster index of each data record. 

Afterwards, for each road segment, we extract a $10 \times 3$ feature matrix $V$ as follows. Suppose there are $n$ historical records within the $i$th cluster, while for this road segment, the numbers of congestion labels for the undefined/green, yellow, and red states are $c_i^1$, $c_i^2$, and $c_i^3$, respectively. Then the $i$th row of $V$ is calculated as $(c_i^1, c_i^2, c_i^3)/n$. 
In this way, we obtain the statistical distribution of the congestion level of each road segment w.r.t. different global traffic volumes, which are taken as the prior knowledge and passed into the graph neural network for further learning.

\subsection{Volume Feature Learning}
In our road graph, loop counters are very sparse. If the volumes of loop counters are directly introduced into the network as node features, there will be many vacant values in nodes. So we use a multi-layer perception network to learn the relationship between counter volume and road segments, and then feed the feature of road segments to the graph neural network for further learning.
\subsection{Static Feature Learning}
The attribute features of road segments are essential. We embed $\textit{importance}$ to $\vect {R}^{5}$, $\textit{oneway}$ to $\vect {R}^{2}$, $\textit{tunnel}$ to $\vect {R}^{2}$, and $\textit{lanes}$ to $\vect {R}^{3}$. For continuous features like $\textit{parsed max speed}$, \textit{flow speed},  $\textit{length meters}$, $\textit{counter distance}$, and $\textit{limit speed}$ we just concatenate them. By this way, We obtain a unified representation for each road segments.

\begin{figure}[t]
	\centerline{\includegraphics[width=0.8\linewidth]{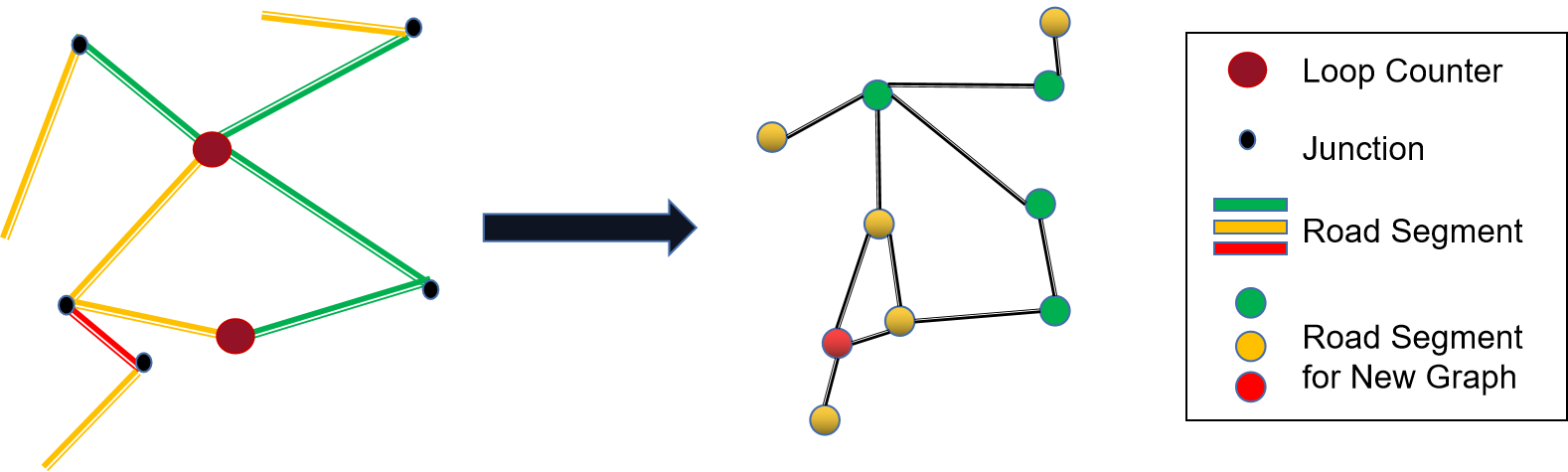}}
	\caption{Constructing the graph by using road segments as nodes.}
	\label{fig:real-2}
\end{figure}

\subsection{Graph Neural Network}
GNN has achieved good results in many fields, it can aggregate information of other nodes and edges through a message passing mechanism. Notice that the loop counter volume data is sparse, which means the nodes are hard to learn the features of their neighbors by normal graph neural network. 
Moreover, our tasks aim to learn the features of road segments, but the GNN model aggregate information based on nodes. So it is more conducive to the representation learning of road segment features by taking road segments as the nodes of the graph. For this reason, we construct a new road graph, taking road segments as the nodes of the graph. In this way, we can better learn the dependencies between road segments. We introduce the features learned by the previous modules into the multi-layer graph neural network, learn the spatial dependency relationship between road segments, and pass the output to the multi-task learning network. 

\subsection{Multi-task Learning}
We propose a multi-task learning component to the learn congestion class, the speed value, and the volume class for each segment in the road graph in a concurrent way. We feed the output of GNN module to residual blocks~\cite{he2016deep} to obtain the congestion class $\vect {y}_{c}^{pred}$, speed $\vect{y}_{s}^{pred}$, volume class $\vect{y}_{v}^{pred}$. The loss function is defined as follows:

\begin{eqnarray} 
L_{c} = Weighted Cross Entropy\big(y_{c_{i}}, y_{c_{i}}^{pred}\big)
\end{eqnarray}

\begin{equation}
L_{s} = \left(y_{s_{i}}-y_{s_{i}}^{pred}\right)^2
\end{equation}

\begin{eqnarray} 
L_{v} = Weighted Cross Entropy\big(y_{v_{i}}, y_{v_{i}}^{pred}\big)
\end{eqnarray}
Where $\vect {y}_{c}$ denotes the label of the congestion class (0=undefined; 1=green/uncongested; 2=yellow/slowed-down; 3=red/congested), $\vect {y}_{s}$ denotes the label of the speed, and $\vect {y}_{v}$ denotes the label of the volume class (1 for volumes 1 and 2; 3 for volumes 3 and 4; 5 for volumes 5 and above, according to the competition).

Finally, we combine the loss terms of these tasks as the overall loss function:

\begin{eqnarray} 
L = \lambda_{1} \cdot L_{c}  + \lambda_{2} \cdot L_{s} + \lambda_{3} \cdot L_{v} 
\end{eqnarray}

where $ \lambda_1 $, $ \lambda_2$, and $ \lambda_3$  are hyper-parameters. We take 0.03, 1, 1 in training.

\subsection{ETA Prediction}
Since we have the speed output and length of all road segments, we just need to calculate the travel time of segments in the super-segments and add them all together to produce the super-segment ETAs.

\section{Experiments}
In this section, we examine the performance of our method on three real-world large-scale datasets from Traffic4cast competition 2022, collected by HERE Technologies in the years 2019 to 2021. There are three city datasets: London, Madrid, and Melbourne. For each city, its road graph consists of tens of thousands of nodes, edges, and some sparse loop counters. For example, the road graph of London has 132,414 edges, 59,110 nodes, and 3,751 counters. Loop counters volume is aggregated every 15 minutes, resulting in 3,751x4 volume input in which the vacant value is filled with 0. In the training set, we only select the data in the daytime, because the loop counters data and label data are sparse at midnight.

\begin{table}[H]
	\centering
	\caption{Datasets}
	\label{tb:performance-comparison-1}
	\begin{tabular}{ccccc}
		\toprule
		Cities& Nodes& Edges &Counters & Period\cr
		\midrule
		London& 59110 & 132414 & 3751  & 2019-07-01 - 2020-01-31\cr
		Madrid& 63397  & 121902 & 3875 & 2021-06-01 - 2021-12-31\cr
		Melbourne&49510 & 94871 & 3982  & 2020-06-01 - 2020-12-30\cr
		\bottomrule
	\end{tabular}
\end{table}

\begin{table}[th]
	\centering
	\caption{Model setting}
	\label{tb:performance-comparison-2}
	\begin{tabular}{cccccc}
		\toprule
		model (checkpoint) & sampling &size training set & number of iterations &batch size &trainable parameters\cr
		\midrule
	model<London>& 110 days at 6am-10pm & 110*64 & 20 epochs  & 2 & 525.03M\cr
	model<madrid>& 109 days at 6am-10pm & 109*64 & 20 epochs  & 2 & 502.60M\cr
        model<melbourne>& 106 days at 6am-10pm & 106*64 & 20 epochs & 2 &409.66M\cr
		\bottomrule
	\end{tabular}
\end{table}

\begin{minipage}{1\textwidth}
    \begin{minipage}[t]{0.5\textwidth}
    \begin{table}[H]
	\centering
	\caption{Leaderboard for core challenge}
	\label{tb:performance-comparison-3}
	\begin{tabular}{cc}
		\toprule
		 Team & core challenge \cr
		\midrule
		 ustc-gobbler & 0.8431079388\cr
		 Bolt& 0.8496679068\cr
		 oahciy & 0.8504153291\cr
		 GongLab & 0.8560309211\cr
		 TSE & 0.8736550411\cr
		 discovery & 0.8759126862\cr
		 ywei & 0.8778917591 \cr
		\bottomrule
	\end{tabular}
\end{table}
    \end{minipage}
\begin{minipage}[t]{0.5\textwidth}
\begin{table}[H]
	\centering
	\caption{Leaderboard for extended challenge}
	\label{tb:performance-comparison-4}
	\begin{tabular}{cc}
		\toprule
		 Team & extended challenge \cr
		\midrule
		 ustc-gobbler  &58.49972153 \cr
		 TSE  & 59.78244781 \cr
		 oahciy  & 61.22274017 \cr
		 Bolt &61.2546107 \cr
		 discovery & 62.29674403 \cr
		 GongLab & 64.74489975 \cr
		\bottomrule
	\end{tabular}
    \end{table}
        \end{minipage}
    \end{minipage}

\subsection{Leaderboard Results}
For each city dataset, we trained 9 models, with each model taking 20 rounds of training in order to select the best-performed one on the validation set. Then, we average the predicted results of the 9 model to produce the final submission result. Finally, our team GongLab won the fourth place in the core competition and the sixth place in the extended competition.

\subsection{Methods for Comparison}
We compared our model against several baselines on the validation set:
\begin{itemize}
    \item Naive Count: For the core challenge, we count the congestion category of all roads and calculate the probability. For the extended challenge, we calculate the median eta of all super-segments in the training set.
    \item Volume Cluster: The baseline groups the data based on the sum of counter volumes. It calculates the statistical probability of congestion for different groups for the core challenge and the median ETA for different groups for the extended challenge.
    \item GNN: The GNN model is provided by the organizers. It fills the loop counter volumes of nodes as input and uses message passing mechanism to learn the dependence between nodes.
\end{itemize}

\begin{table}[t]
	\centering
	\caption{Performance for core challenge on validation set.}
	\label{tb:performance-comparison-5}
	\begin{tabular}{ccccc}
		\toprule
	Models&Naive count&Volume cluster&GNN&OurModel\cr
		\midrule
		London& 1.0013 & 0.9872 & 0.9702& 0.82705\cr
		Madrid& 1.0023 & 0.9914 & 0.9735&0.82652\cr
		Melbourne& 1.0216  & 1.0066 & 0.9845 & 0.84560 \cr
		\bottomrule
	\end{tabular}
\end{table}
\begin{table}[t]
	\centering
	\caption{Performance for extended challenge on validation set.}
	\label{tb:performance-comparison-6}
	\begin{tabular}{ccccc}
		\toprule
	Models&Naive count&Volume cluster&GNN&OurModel\cr
		\midrule
		London& 108.0988 & 99.9642 & - & 91.0117\cr
		Madrid& 68.1231 & 61.2911 & - & 59.8746\cr
		Melbourne& 41.2249 & 37.8772 & - & 39.1628\cr
		\bottomrule
	\end{tabular}
\end{table}

As can be seen from Table 5 and Table 6, our model performance is significantly better than the baselines. The Naive Count model does not take the dynamic changes of input into account, and Volume Cluster model cannot learn the dependence between road segments. Since the loop counter volumes are very sparse, the performance of the GNN model is poor. Our model combines the Volume Cluster and GNN methods, which can capture the dynamics of volume data as well as the dependencies between road segments. Moreover, we adopt the multi-task learning method to learn the congestion class, the speed value, and the volume class simultaneously, which further boosts the performance. 

\begin{table}[H]
	\centering
	\caption{Performance Comparison of our model and the ablation experimental model.}
	\label{tb:performance-comparison-7}
	\begin{tabular}{ccccc}
		\toprule
	Models&OurModel&No cluster&No static feature&No GNN\cr
		\midrule
		London& 0.82705 & 0.84949 &0.83534 & 0.82910\cr
		Madrid& 0.82652 &0.84946 &0.83442 & 0.82825\cr
		Melbourne& 0.84560  & 0.89053 & 0.84874 & 0.84803\cr
		\bottomrule
	\end{tabular}
\end{table}
We perform ablation experiments to verify the effectiveness of the model. It can be seen from Table 7 that the performance reduces after removing each module we proposed, which shows that the above modules are meaningful. For example, if the clustering component is removed, the score of the model increases from 0.82705 to 0.84949, indicating that it is important to use volume clustering method to capture the dynamic changes of loop counters data.

\section{Discussion}
We propose a multi-task learning framework for sparse traffic prediction, and experimental results show that our model is significantly better than the baselines. During the experiments, we also found that for particularly sparse node input, it is difficult for the general GNN network to learn the spatial dependence between nodes. 
Besides, taking the clustering technique to analyze the historical data as additional patterns can effectively improve the performance of the model.

\section{Acknowledgement}
This work was supported in part by the National Natural Science Foundation of China under Grant 62276100, in part by the Guangdong Natural Science Funds for Distinguished Young Scholars under Grant 2022B1515020049, and in part by the Guangdong Regional Joint Funds for Basic and Applied Research under Grant 2021B1515120078. 

\bibliographystyle{unsrt}
\bibliography{references}



\end{document}